\documentclass[letterpaper]{article} 
\usepackage{aaai25}  
\usepackage{times}  
\usepackage{helvet}  
\usepackage{courier}  
\usepackage[hyphens]{url}  
\usepackage{graphicx} 
\usepackage{natbib}  
\usepackage{caption} 
\usepackage{enumitem} 
\usepackage{graphicx}
\usepackage{bm} 
\usepackage{times}
\usepackage{amsmath}
\usepackage{amsthm}
\usepackage{amsfonts}
\usepackage{multirow} 
\usepackage{booktabs}
\usepackage{xcolor}
\usepackage{colortbl} 
\usepackage[ruled,vlined,linesnumbered,lined,boxed,commentsnumbered]{algorithm2e}
\usepackage{algpseudocode}
\usepackage{caption}
\usepackage{subfigure}

\title{CoDTS: Enhancing Sparsely Supervised Collaborative Perception \\ with a Dual Teacher-Student Framework}
\author{
    Yushan Han, Hui Zhang, Honglei Zhang, Jing Wang, Yidong Li\thanks{Corresponding author (ydli@bjtu.edu.cn).}\\
}
\affiliations{
    Key Laboratory of Big Data \& Artificial Intelligence in Transportation (Beijing Jiaotong University), Ministry of Education\\
    School of Computer Science and Technology, Beijing Jiaotong University, Beijing 100044, China\\

    \{21112034, huizhang1, honglei.zhang, jing\_w, ydli\}@bjtu.edu.cn
}

\usepackage{bibentry}

\begin{document}

\maketitle

\begin{abstract}
    Current collaborative perception methods often rely on fully annotated datasets, which can be expensive to obtain in practical situations. 
    To reduce annotation costs, some works adopt sparsely supervised learning techniques and generate pseudo labels for the missing instances. 
    However, these methods fail to achieve an optimal confidence threshold that harmonizes the quality and quantity of pseudo labels.
      To address this issue, we propose an end-to-end \textbf{Co}llaborative perception \textbf{D}ual \textbf{T}eacher-\textbf{S}tudent framework (CoDTS), which employs adaptive complementary learning to produce both high-quality and high-quantity pseudo labels. 
      Specifically, the Main Foreground Mining (MFM) module generates high-quality pseudo labels based on the prediction of the static teacher. Subsequently, the Supplement Foreground Mining (SFM) module ensures a balance between the quality and quantity of pseudo labels by adaptively identifying missing instances based on the prediction of the dynamic teacher. Additionally, the Neighbor Anchor Sampling (NAS) module is incorporated to enhance the representation of pseudo labels. To promote the adaptive complementary learning, we implement a staged training strategy that trains the student and dynamic teacher in a mutually beneficial manner.
      Extensive experiments demonstrate that the CoDTS effectively ensures an optimal balance of pseudo labels in both quality and quantity, establishing a new state-of-the-art in sparsely supervised collaborative perception.
      The code is available at \url{https://github.com/CatOneTwo/CoDTS}.
\end{abstract}

%

\section{Introduction}


Multi-agent collaborative perception \cite{han2023collaborative}  can enhance the perception capabilities of vehicles in autonomous driving scenarios, such as vehicle-to-vehicle (V2V), vehicle-to-infrastructure (V2I), and vehicle-to-everything (V2X) interactions. As an emerging research, recent works have contributed valuable datasets, efficient communication mechanisms, and effective fusion techniques. Additionally, several pioneer approaches aim to address practical challenges like localization errors, communication latency and model heterogeneity.

However, classic collaborative perception methods often depend on fully annotated datasets. Despite their considerable performance, they are challenged by being labor-intensive and time-consuming.
For example, in the case of LiDAR-based 3D object detection, it takes more than 100 seconds \cite{song2015sun,meng2020ws3d,luo2023exploring} to annotate a precise 3D bounding box for individual perception datasets such as KITTI \cite{geiger2012we} and Waymo \cite{sun2020scalability}. Since collaborative perception involves multiple agents, the annotation cost becomes several times higher than that of individual perception. 
Moreover, the annotation of collaborative perception requires aligning shared objects across multiple agents, which increases annotation complexity.
Therefore, the data annotation significantly impedes the practical application of collaborative perception.


To address the above high-cost issue, an alternative approach is sparsely supervised learning, where only one object per agent in each frame is randomly annotated, as shown in Figure~\ref{fig1}(b).
To mine missing instances, the current mainstream methods \cite{han2023ssc3od,HINTED} primarily adopt a teacher-student framework. Specifically, SSC3OD \cite{han2023ssc3od} proposes a self-supervised learning approach to pre-train the teacher model. Then, it utilizes a static teacher-student framework to produce high-quality pseudo labels using high thresholds.
HINTED \cite{HINTED} adopts a dynamic teacher-student framework, where the teacher mainly relies on high-level pseudo labels and a mixed-density student is incorporated to ensure that the model prioritizes hard instances. While these methods improve performance to a certain degree, the detection performance under sparse supervision still lags significantly behind that of full supervision. 
We argue that these methods primarily focus on the quality of pseudo labels, overlooking their quantity, thus resulting in suboptimal solutions.

\begin{figure}[htbp]
  \centerline{\includegraphics[width=1\linewidth]{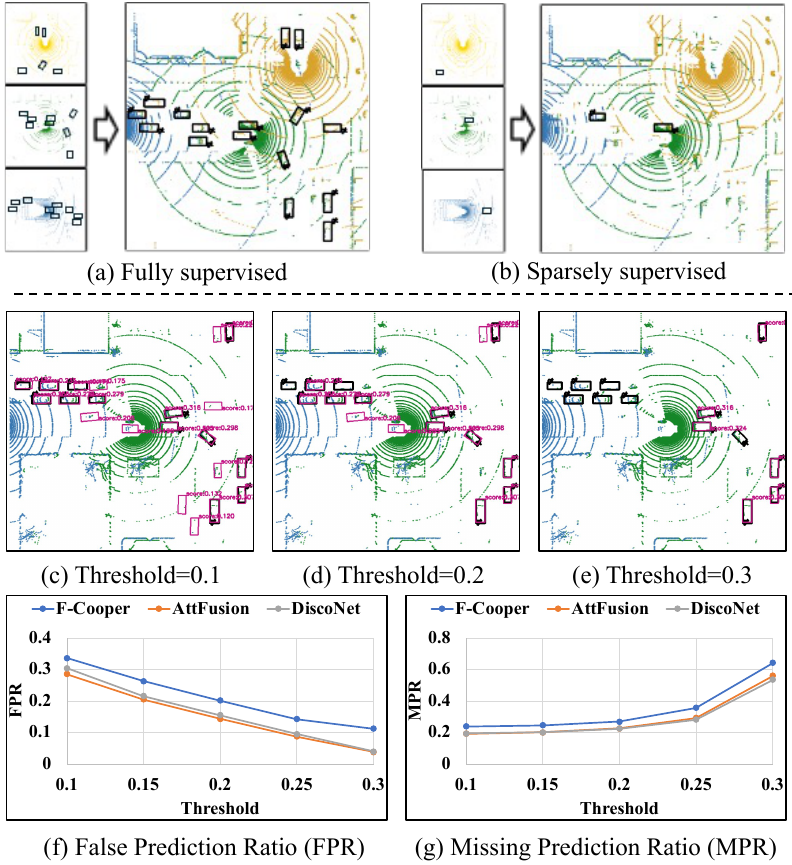}}
  \caption{(a)-(b): Labels of fully and sparsely supervised collaborative perception. (c)-(g): The impact of different confidence thresholds on \textcolor[RGB]{199,21,133}{pseudo labels}. The false prediction ratio (FPR) = the number of false predictions / the number of pseudo labels, and the missing prediction ratio (MPR) = the number of missed predictions / the number of ground truths.}
  \label{fig1}
  \vspace{-1.5em}
\end{figure}



Specifically, traditional teacher-student frameworks often employ higher thresholds to derive high-quality pseudo labels. However, varying threshold levels present unique benefits. As illustrated in Figure~\ref{fig1} (c)-(g), a lower threshold may capture the majority of foreground objects (low MPR) but at the expense of introducing substantial background noises (high FPR). In contrast, a higher threshold can ensure that the selection primarily includes foreground objects (low FPR) but may result in the loss of important foreground details (high MPR). Consequently, \textit{how to find an optimal confidence threshold that balances the quality and quantity of pseudo labels remains a formidable challenge}.

To tackle this challenge, we propose an end-to-end collaborative perception dual teacher-student framework (CoDTS), which leverages adaptive complementary learning to generate pseudo labels that are both high in quality and quantity, ensuring optimal performance.
This is achieved through a staged training strategy: i) In the warm-up stage, the Main Foreground Mining (MFM) module is introduced to ensure a sufficient quantity of pseudo labels. This mainly involves filtering the prediction of the static teacher using a low confidence threshold, which effectively pre-trains the student and dynamic teacher. ii) In the refinement stage, the MFM focuses on identifying high-quality positive instances by adjusting the confidence threshold of the static teacher to a higher value, which inevitably causes missing positive instances. To complement these instances, we introduce Supplement Foreground Mining (SFM) module, which identifies the missing ones by filtering the predictions of dynamic teachers with a dynamically adapted high confidence threshold. This adaptive complementary learning guarantees an optimal balance of both quality and quantity pseudo labels. Furthermore, during both stages, the Neighbor Anchor Sampling (NAS) module is used to select neighboring instances for the identified positives, enhancing the representations of pseudo labels.
Besides, the framework trains the student and dynamic teacher through mutual learning, enabling them to make continuous improvements.
Our contributions can be summarized as follows:

\begin{itemize}[itemsep=0pt,topsep=3pt,parsep=4pt]
  \item We propose a novel dual teacher-student framework CoDTS, which effectively resolves the imbalance of pseudo labels between quality and quantity in traditional teacher-student frameworks. It enables sparsely supervised collaborative detectors to achieve comparable performance to fully supervised ones.
  \item The proposed MFM and SFM are used to identify both high-quality and high-quantity pseudo labels through adaptive complementary learning, with the threshold in SFM dynamically adapted according to current scenes. Besides, the NAS is employed to enhance the dense representation of pseudo labels.
  \item A staged training strategy is introduced to perform adaptive complementary learning in an end-to-end manner. By implementing this strategy, the student and dynamic teacher are optimized in a mutually beneficial manner, enabling them to evolve together and continuously enhance detection accuracy.
  \item Extensive experiments are conducted on V2X-Sim, OPV2V, DAIR-V2X and V2V4Real datasets to demonstrate the superior performance of CoDTS framework.
\end{itemize}

\begin{figure*}[htbp]
  \centerline{\includegraphics[width=1\linewidth]{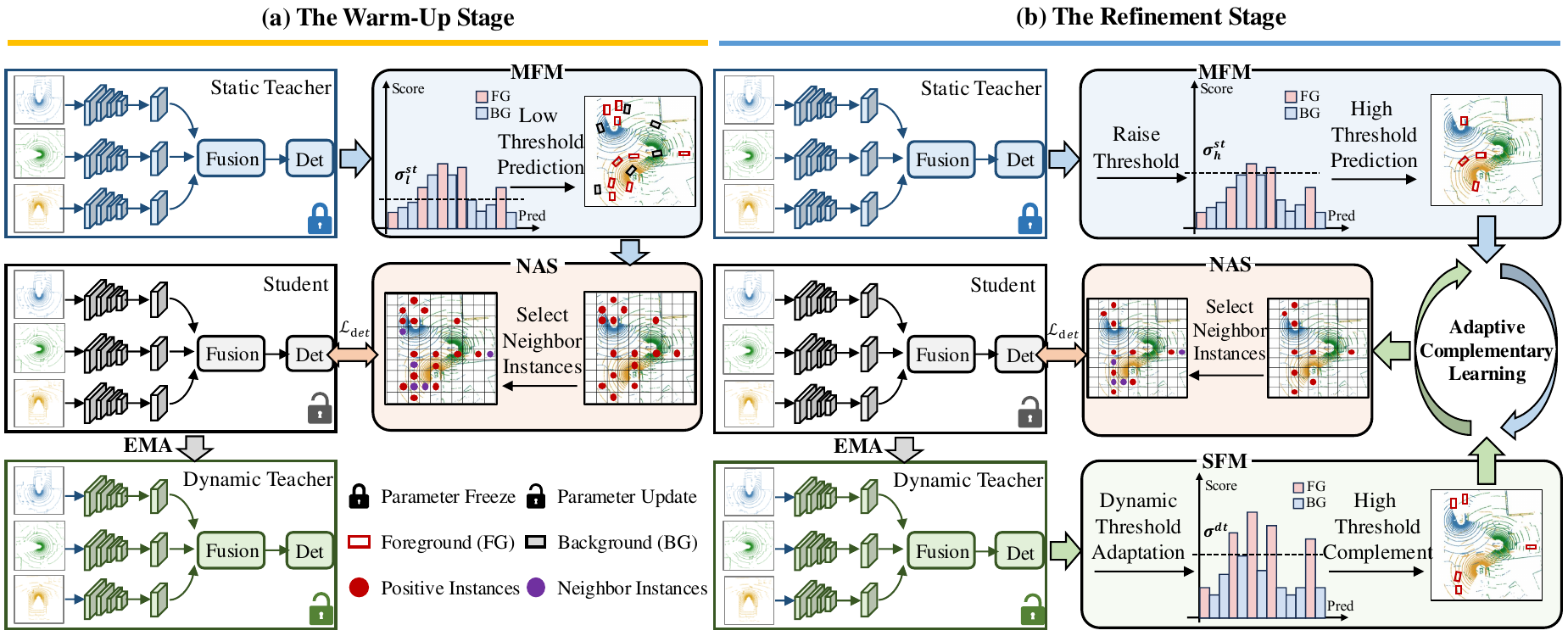}}
  \caption{ 
    The CoDTS framework employs a staged training strategy. (a) In the warm-up stage, the MFM utilizes a low threshold to generate a sufficient amount of pseudo labels, which are used to guide the student and pre-train the dynamic teacher. (b) In the refinement stage, the MFM raises the threshold to obtain high-quality pseudo labels, while the SFM utilizes dynamic threshold adaptation to adjust the threshold and adaptively identify missing instances to complement the MFM. This Adaptive Complementary Learning ensures the generation of \textit{both high-quality and high-quantity} pseudo labels, which are then merged to guide the student. Throughout both stages, the NAS is used to enhance the representations of pseudo labels.
    }
  \label{fig2}
  \vspace{-1em}
\end{figure*}

\section{Related Work}

\subsection{Collaborative Perception}
Collaborative perception enhances the individual perception \cite{zhang2022parallel,luo2023one} by facilitating information exchange among multiple agents, enabling them to obtain more comprehensive views. Previous research has achieved notable advancements in effective communication \cite{where2comm,yang2023how2comm} and adaptive fusion \cite{V2X-VIT,CRCNet,su2024makes} under ideal conditions and in addressing challenges like time delay \cite{yu2023flow} and localization errors \cite{ERMVP,MRCNet} in real-world applications. 
However, these methods heavily rely on complete annotations, which are labor-intensive and impede the practical deployment of collaborative perception.

\subsection{Sparsely Supervised 3D Object Detection}

To mitigate the high annotation costs in LiDAR-based 3D object detection, researchers have turned to sparsely supervised approaches. SS3D \cite{liu2022ss3d} pioneers sparsely supervised 3D object detection, iteratively mining and storing reliable instances and backgrounds under sparse supervision to ultimately achieve full supervision. SSC3OD \cite{han2023ssc3od} first introduces sparsely supervised collaborative 3D object detection, utilizing a Pillar-based mask autoencoder to aid the detector in reasoning about high-level semantics. HINTED \cite{HINTED} employs a dynamic teacher-student framework and proposes a mixed-density student to boost performance in hard instances.  Despite significant advances in single-vehicle 3D object detection, sparsely supervised collaborative 3D object detection has yet to achieve satisfactory performance. This work seeks to enhance the performance of collaborative 3D object detectors under sparse supervision.

\section{Methodology}

\subsection{Overview Framework}

To address the imbalance of pseudo labels between quality and quantity, we propose the CoDTS framework, which generates pseudo labels through adaptive complementary learning. Specifically, the MFM is proposed to generate high-quality pseudo labels based on the prediction of the static teacher, which inevitably misses some instances. To complement the MFM, the SFM is introduced to adaptively identify the missing instances based on the prediction of the dynamic teacher, ensuring the quality and quantity of pseudo labels. Additionally, the NAS is used to enhance the dense representations of pseudo labels during training by selecting neighboring instances. 
As shown in Figure~\ref{fig2}, the proposed CoDTS framework employs a delicate staged training strategy. In the warm-up stage, the student and dynamic teacher are effectively pre-trained. In the refinement stage,  we implement adaptive complementary learning to produce pseudo labels. Furthermore, the dynamic teacher and student are jointly trained in a mutually beneficial manner, enabling continuous optimization. Next, we will delve into the details of adaptive complementary learning, neighbor anchor sampling, and staged training strategy.

\subsection{Adaptive Complementary Learning} 
To generate both high-quality and high-quantity pseudo labels, we introduce the adaptive complementary learning regimen. Specifically, the MFM generates pseudo labels by using a high threshold to filter the prediction of the static teacher. Following this, the SFM adaptively complements the missing pseudo labels from MFM, with thresholds dynamically adjusting to the current scenario.

In a collaborative detector, each fused feature generates anchors $R\in \mathbb{R}^{H \times W \times A}$, where $H$ and $W$ represent the dimensions of the feature map, and $A$ represents the number of anchors in each feature grid. The detection prediction consists of three types of information, i.e., $z=\{z_{cls}, z_{reg}, z_{dir}\}$, where $z_{cls}$, $z_{reg}$ and $z_{dir}$ represent the classification, regression and direction for each anchor, respectively. The details of MFM and SFM are as follows.



\subsubsection{Main Foreground Mining} 
Given the static teacher $\theta^{st}$, confidence threshold $\sigma^{st}$ and overlap threshold $\tau$, the function of identifying positive instances $R^{st}$ by MFM can be denoted as follows:
\begin{equation}
  R^{st}=\bold{MFM}(\theta^{st}, \sigma^{st}, \tau). \label{MFM}
\end{equation}

Specifically, the MFM first selects anchors $R_{0}^{st}$ whose classification prediction $z_{cls}^{st}$ is larger than confidence threshold $\sigma^{st}$. Subsequently, it utilizes the Non-Maximum Suppression (NMS) algorithm \cite{nms} with an overlap threshold $\tau$ to eliminate anchors with significant overlap. Finally, the remaining anchors are treated as positive instances $R^{st}$. This process can be formulated as follows:
\begin{align}
  R_{0}^{st}&=\{ R^{(i)} \mid z_{cls}^{st(i)} > \sigma^{st},R^{(i)}\in R\}, \nonumber \\ 
  R^{st}&=\text{NMS}(R_{0}^{st},\tau). 
\end{align}

\subsubsection{Supplement Foreground Mining} 
The SFM aims to identify missing positive instances using the dynamic teacher $\theta^{dt}$. Similar to MFM, the SFM also utilizes a confidence threshold $\sigma^{dt}$ and an overlap threshold $\tau$ to identify instances $R^{dt}$ with high confidence and low overlap, which can be denoted as:
\begin{equation}
  R^{dt}=\bold{SFM}(\theta^{dt}, \sigma^{dt}, \tau). \label{SFM}
\end{equation}

\begin{figure}[htbp]
  \centerline{\includegraphics[width=1\linewidth]{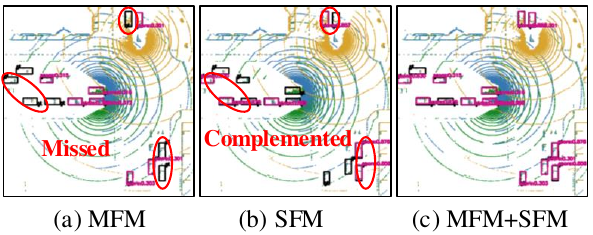}}
  \caption{Visualization of \textcolor[RGB]{199,21,133}{pseudo label} generation through adaptive complementary learning. (a)-(c) represent the pseudo labels generated by MFM, SFM, and their merged results, respectively.}
  \label{fig5}
  \vspace{-1em}
\end{figure}

The dynamic teacher $\theta^{dt}$ exhibits varying detection performance across different scenes. For instance, it might predict high scores in some scenes but low scores in others, making it unreasonable to use a predefined static threshold. To ensure the dynamic teacher threshold $\sigma^{dt}$ seamlessly adapts to diverse scenarios and effectively filters high-quality foregrounds, we employ a \textbf{dynamic threshold adaptation} method, eliminating the need for human intervention.


Specifically, we collect the predicted scores of the dynamic teacher for sparse labels in the current batch and apply the sigmoid function to generate classification scores. Since the sparse labels provide accurate supervision, these scores should generally be at a similar level, though some outliers may have lower scores. We use the K-Means clustering algorithm \cite{kmeans} to classify these scores into high-level and low-level clusters. The high-level cluster center, representing the average prediction level of the dynamic teacher $\theta^{dt}$ in the current batch, is chosen as threshold $\sigma^{dt}$.

Since we pre-define $A$ anchors for each feature grid, both the static and dynamic teachers generate predictions for each anchor. Consequently, different teachers may have different predictions for the same grid. To avoid conflicting predictions, we calculate the feature grid index for positive instances in $R^{dt}_1$ and $R^{st}$, and only remain the instances in $R^{dt}_1$ that are missed by $R^{st}$.
We denote the index function of anchors as $\text{Id}(\cdot)$, and the overall process of SFM can be formulated as follows:
\begin{align}
  R_{0}^{dt}&=\{ R^{(i)} \mid z_{cls}^{dt(i)} > \sigma^{dt},R^{(i)}\in R\}, \nonumber \\
  R_{1}^{dt}&=\text{NMS}(R_{0}^{dt},\tau),\nonumber \\
  R^{dt}&=\{ R^{(i)} \mid \text{Id}(R^{(i)})/A \notin \text{Id}(R^{st})/A, R^{(i)}\in R_{1}^{dt}\}. 
\end{align}


Figure~\ref{fig5} illustrates the pseudo labels generated via adaptive complementary learning. The pseudo labels generated by MFM are all positive instances but miss many ground truths. Meanwhile, the pseudo labels generated by SFM effectively complement these missed instances. As a result, the merged pseudo labels from MFM and SFM encompass nearly all ground truths while maintaining high quality.

\begin{figure}[htp]
  \centerline{\includegraphics[width=1\linewidth]{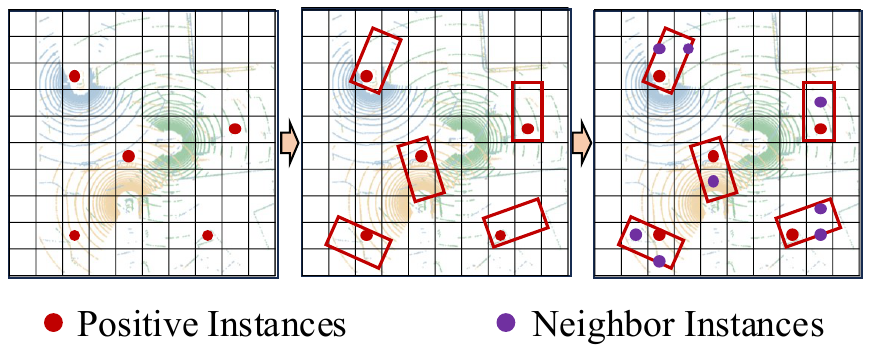}}
  \caption{The process of NAS. First, we generate bounding boxes for positive instances. We then select neighbor instances that have a high overlap with these bounding boxes. Finally, the positive instances are combined with the neighbor instances to create dense positive instances.}
  \label{fig3}
  \vspace{-1em}
\end{figure}

\subsection{Neighbor Anchor Sampling} 
While MFM and SFM effectively identify positive instances, the mined instances are too sparse compared to the entire feature grid, which makes it challenging to learn a foreground representation. Inspired by the fact that highly spatially overlapped instances should have the same label, we propose the NAS, which selects the neighbor instances based on these identified positive instances, thereby enhancing the dense representation of pseudo labels.

As shown in Figure~\ref{fig3}, consider the positive instances $R^{st}$ (represented as red dots) identified by MFM. We use the regression prediction $z_{reg}^{st}$ of instances to calculate their corresponding bounding boxes $B^{st}$ (represented as red rectangles). Subsequently, we compute the overlaps between all anchors $R$ and the foreground bounding boxes $B^{st}$ and set an overlap threshold $\tau^{nei}$ to select a set of neighbor instances $R^{nei}$ (represented as purple dots). The above procedure can be deﬁned as:
\begin{align}
  R^{nei}&=\bold{NAS}(R^{st}, \tau^{nei}) \nonumber \\
  &=\{ R^{(i)} \mid \text{IoU}(R^{(i)},B^{st}) > \tau^{nei}, R^{(i)}\in R\}. \label{NAS} 
\end{align}

By utilizing NAS, we enhance the representation of pseudo labels, which assists collaborative detectors in learning these instances more effectively. Finally, we will combine the sparse labels with the pseudo labels to generate the ultimate pseudo label. If both the sparse and pseudo labels identify the same anchor as a positive instance, we retain only the sparse label to ensure supervision accuracy.

\subsection{Staged Training Strategy} \label{A}

To implement adaptive complementary learning, our CoDTS employs a delicate staged training strategy. As shown in Algorithm~\ref{STT}, the staged training strategy includes two phases: the warm-up stage and the refinement stage.

\subsubsection{The Warm-Up Stage}
The purpose of the warm-up stage is to generate \textit{high-quantity} pseudo labels to guide the student and pre-train the dynamic teacher, which is used for the generation of complementary pseudo labels in the subsequent stage. First, we use the MFM to identify positive instances by applying a low threshold $\sigma^{st}_l$ to filter the predictions of the static teacher. This ensures that the pseudo labels include sufficient foreground information. Next, we use the NAS to select neighboring instances, enhancing the dense representation of the pseudo labels.

After each training iteration $iter$, the student $\theta^s$ utilizes the exponential moving average (EMA) with a smoothing coefficient $\alpha$ to update the parameter weight of the dynamic teacher $\theta^{dt}$. By the end of the warm-up stage, both the student and the dynamic teacher can effectively identify most foregrounds. The process of EMA is defined as follows:
\begin{equation}
  \theta_{iter}^{dt}= \begin{cases}\left(1-\frac{1}{iter}\right) \times \theta_{iter-1}^{dt}+\frac{1}{iter} \times \theta^s_{iter}, & 1-\frac{1}{iter}<\alpha \\ 
  \alpha \theta_{iter-1}^{dt}+(1-\alpha) \theta^s_{iter}. & \text {otherwise}\end{cases}
  \label{ema}
  \end{equation}

  By employing the EMA strategy, we can view the slowly evolving dynamic teacher as an ensemble of the students at various iterations, which enhances the stability and robustness of the prediction capabilities of the dynamic teacher. Additionally, positive instance features remain consistent throughout training, while negative instance features exhibit random diversity. Consequently, the dynamic teacher will assign higher confidence to positive instances, which sets the foundation for utilizing the dynamic teacher to identify missing positive instances in the subsequent stage.

\subsubsection{The Refinement Stage}

The refinement stage aims to generate \textit{both high-quality and high-quantity} pseudo labels to improve the detection accuracy of both student and dynamic teacher. This is accomplished through adaptive complementary learning between MFM and SFM. Specifically, we first raise the threshold of MFM to a high level $\sigma^{st}_h$ to identify positive instances, ensuring the quality of pseudo labels but inevitably missing some positive instances. Then the SFM applies a high threshold $\sigma^{dt}$ for SFM to complement the missing instances adaptively, where $\sigma^{dt}$ is dynamically adapted based on current scenes. This complementary mining approach ensures an optimal balance between the quality and quantity of pseudo labels.
Subsequently, we merge the positive instances identified by both MFM and SFM, and use NAS to select neighbor instances. Similar to the warm-up stage, we use EMA strategy to update the dynamic teacher.

In the refinement stage, the student and dynamic teacher are optimized via mutual learning, where the dynamic teacher generates part of high-quality pseudo labels to train the student, and the student updates the knowledge it learned back to the dynamic teacher.
With the above interaction, both models can evolve jointly and continuously to improve detection accuracy, which also means that the dynamic teacher can generate more accurate and stable pseudo labels.
Furthermore, due to continuous supervision by pseudo labels, both student and dynamic teacher are able to detect a large number of high-quality objects effectively.

\begin{algorithm}
  \SetKwData{Left}{left}\SetKwData{This}{this}\SetKwData{Up}{up}
  \SetKwFunction{Union}{Union}\SetKwFunction{FindCompress}{FindCompress}
  \SetKwInOut{Input}{input}\SetKwInOut{Output}{output}

  \Input{Static teacher $\theta^{st}$. Dynamic teacher $\theta^{dt}$. Student $\theta^{s}$.  Refinement starts iteration $I_{refine}$. Max training iteration $I_{max}$}
  
  \textbf{Preparation:} \\
  Initialize the weight of student model $\theta^{s}$ \\
  Copy the weight of student $\theta^{s}$ to dynamic teacher $\theta^{dt}$ \\
  \textbf{Do training iteration:}\\
  \For{{$iter = 1,2,...,I_{max}$}}{ 
    \BlankLine
    \If(\Comment{\emph{The Warm-Up Stage}}){$iter < I_{refine}$}{
      $R^{st}=\bold{MFM}(\theta^{st}, \sigma^{st}_l, \tau)$ \\
      $R^{nei}=\bold{NAS}(R^{st}, \tau^{nei})$\\
    }
    \BlankLine
    \Else(\Comment{\emph{The Refinement Stage}}){
      $R^{st}=\bold{MFM}(\theta^{st}, \sigma^{st}_h, \tau)$ \\
      Get $\sigma^{dt}$ by dynamic threshold adaptation \\
      $R^{dt}=\bold{SFM}(\theta^{dt}, \sigma^{dt}, \tau)$ \\
      $R^{nei}=\bold{NAS}(R^{st} \cup R^{dt}, \tau^{nei})$ \\
    }
    Merge pseudo labels and sparse labels\\
    Calculate loss and update the weight of $\theta^{s}$\\
    Update the weight of $\theta^{dt}$ by EMA
    }
    \Output{Student $\theta^{s}$ and dynamic teacher $\theta^{dt}$}
  \caption{Staged Training Strategy}\label{STT}
\end{algorithm}




\begin{table*}[htbp]
  \huge
  \renewcommand\arraystretch{1.2} 
  \centering 
  \resizebox{1\linewidth}{!}{
  \begin{tabular}{c|c|c|ccc|ccc|ccc|ccc}
  \toprule 
  \multicolumn{3}{c}{Datasets}&\multicolumn{3}{c}{V2X-Sim}&\multicolumn{3}{c}{OPV2V}&\multicolumn{3}{c}{DAIR-V2X}&\multicolumn{3}{c}{V2V4Real}\\
  \midrule
  Detectors& Label & Methods & AP@0.3 &AP@0.5 &AP@0.7& AP@0.3 &AP@0.5 &AP@0.7& AP@0.3 &AP@0.5 &AP@0.7 & AP@0.3 &AP@0.5 &AP@0.7\\
  \midrule
  \multirow{6}*{F-Cooper}   & Full & -& 79.69 & 71.59 & 56.75 & 94.23& 88.44& 69.16 & 80.11 & 73.37 & 56.51 & 81.35  &  72.10  & 40.73  \\
   & Sparse & - & 62.62 & 52.14 & 38.72 & 80.83&74.75&51.96 & 54.41 & 49.49 & 32.58  & 46.61  &  43.91  &  23.09 \\
   & Sparse & ST &  67.49 & 60.31 & 45.60 & 82.32 & 76.89 & 54.26 & 59.68 & 54.26 & 35.39 & 53.62 & 50.70 & 30.51 \\
  & Sparse &SSC3OD&  71.20  &  62.49 &  50.09 & 84.40& 79.08 & 57.54 & 60.89 & 56.68 & 40.53 &  55.74 & 52.46 &  33.22  \\
  & Sparse &HINTED& 71.08  & 61.15   &  46.48  & 87.23 & 80.41  & 55.10  &  70.68 & 62.86  & 38.68  &  71.90 & 61.72  & 35.37  \\
  & Sparse &\bf{CoDTS}& \bf{73.77}  & \bf{65.52} & \bf{53.02} & \bf{90.83} & \bf{84.05} & \bf{62.04} & \bf{78.17} & \bf{70.35}& \bf{50.53}  & \bf{71.91}  & \bf{65.60}   & \bf{38.03}  \\
  \midrule
  \multirow{6}*{AttFusion}  & Full & - &79.07 & 76.47 & 66.43 &95.55&93.49&83.03 & 79.82 & 72.89 & 57.43 &  79.81 &  70.37  &  45.20 \\
    & Sparse & -&67.02 & 65.53 & 56.53 &78.74&77.63&61.98 & 56.31 & 51.21 & 32.93 & 48.66  &  46.02  &  26.09 \\
    & Sparse & ST & 69.69 & 68.41 & 59.06 & 80.12 & 79.19 & 68.86 & 57.32 & 52.94 & 38.05 & 55.06 & 50.66 & 33.56  \\
    & Sparse & SSC3OD & 74.10  & 72.68 & 62.66 & 82.35 & 81.64 & 72.55 & 62.02 & 57.17& 40.57 & 56.75 & 53.88   & 35.52  \\
    & Sparse &HINTED&   74.42  &  72.07  & 60.37  & 86.24 & 81.50  & 60.76  & 71.59  &  65.42 & 42.37 & 69.74  & 61.67  & 34.77  \\
    & Sparse & \bf{CoDTS}& \bf{77.84} & \bf{74.86} & \bf{65.66} & \bf{91.23} &\bf{89.42} & \bf{75.77} & \bf{76.95} & \bf{69.23} & \bf{51.24}  & \bf{78.09} &  \bf{69.30}  & \bf{41.30} \\
  \midrule
  \multirow{6}*{DiscoNet} & Full & -& 81.02 & 77.63 & 69.29 & 96.22&94.31 & 84.42 & 80.21 & 73.62 & 58.00 &  79.25 &  68.61  & 42.61  \\
  & Sparse & -&68.56 & 65.39 & 55.84 &81.82&80.79&69.23 & 55.90  & 51.39 & 32.97 & 54.15  & 50.22   & 33.23  \\
  & Sparse & ST &  70.97 & 68.21 & 59.06 & 82.08 & 81.19 & 72.33 & 58.10 & 53.94 & 38.93 & 56.41 & 51.87 & 31.44  \\
   & Sparse & SSC3OD & 74.07 &  70.54 &62.29 & 83.12 & 82.22 & 73.56 &  60.93 &  56.78 &  42.63 & 53.66  & 49.44   & 30.44  \\
   & Sparse & HINTED&  77.12  & 73.18   &  60.60  & 87.20 & 83.24  &  69.61 & 70.19  & 63.78  & 41.00  & 74.87  & 65.34  & 37.95   \\
   & Sparse & \bf{CoDTS}& \bf{81.23} & \bf{76.41} &\bf{66.18} & \bf{93.41} & \bf{90.56} & \bf{76.47} &\bf{75.82} &\bf{68.73} & \bf{51.05} &  \bf{80.45} & \bf{70.56}  & \bf{38.55}  \\
  \bottomrule
  \end{tabular}
  }
  \caption{Performance comparison among different methods on sparsely supervised collaborative 3D object detection.}
  \label{table:performance}
  \vspace{-0.5em}
\end{table*}

  \begin{table}[htbp]
    \normalsize
    \renewcommand\arraystretch{1.1} 
    \resizebox{1\linewidth}{!}{
    \begin{tabular}{c|c|cc|cc}
      \toprule
    \multirow{2}{*}{Detectors} & \multirow{2}{*}{Methods} & \multicolumn{2}{c}{Semi-10\%}     & \multicolumn{2}{c}{Semi-20\%}                    \\
    \cline{3-6}
                               &                                 & AP@0.5         & AP@0.7               & AP@0.5         & AP@0.7         \\
                               \midrule
    \multirow{4}{*}{F-Cooper}   & 3DIoUMatch             & 56.81          & 28.83           & 66.86          & 45.06          \\
                               & HSSDA             &      58.22    &    30.41      &     66.99           &   46.54             \\
                               & HINTED              &     58.93        &    30.42   &   67.15             &      46.22          \\
                               & \textbf{CoDTS}   & \textbf{59.54} & \textbf{32.82}  & \textbf{69.31} & \textbf{48.46} \\
    \midrule
    \multirow{4}{*}{AttFusion}        & 3DIoUMatch                & 54.56          & 31.64          & 63.86          & 45.91          \\
                               & HSSDA         &       57.09         &    34.38      &     63.95           &    47.56            \\
                               & HINTED         &     57.53           &   34.88     &    63.76         &    47.19         \\
                               & \textbf{CoDTS}   & \textbf{58.89} & \textbf{37.26}  & \textbf{66.22} & \textbf{50.23} \\
                               \midrule
    \multirow{4}{*}{DiscoNet}                         & 3DIoUMatch             & 58.78          & 34.88         & 65.96          & 46.81          \\
                               & HSSDA         &      59.51          &    36.56     &       66.45         &     48.28           \\
                               & HINTED      &     60.21           &       \textbf{38.18}   &   66.39          &    48.56            \\
                               & \textbf{CoDTS}  & \textbf{60.61} & 37.94 & \textbf{67.31} & \textbf{49.37} \\
                               \bottomrule
    \end{tabular}}
    \caption{Comparison with state-of-the-art semi-supervised methods on DAIR-V2X. All methods are trained on randomly selected 10$\%$ or 20$\%$ fully annotated frames.}
    \label{table:semi-supervised}
    \vspace{-0.5em}
    \end{table}

\section{Experimental Results}
\subsection{Datasets and Evaluation Metrics}
We conduct experiments on four large-scale collaborative perception datasets and evaluate the methods with average precision (AP) at IoU thresholds of 0.3, 0.5 and 0.7. 

\textbf{V2X-Sim} \cite{V2X-Sim} is a simulated dataset for V2X collaborative perception.
We utilize the V2XSim 2.0 version and set the LiDAR range as $x\in[-32m, 32m], y\in[-32m,32m]$. The training set comprises 698,991 fully supervised objects and 29,300 sparsely supervised objects. 
\textbf{OPV2V} \cite{OPV2V} is a simulated dataset for V2V collaborative perception. 
The LiDAR range is set as $x\in[-140.8m,140.8m], y\in[-38.4m,38.4m]$. 
The training set includes 358,142 fully supervised objects and 21,139 sparsely supervised objects. 
\textbf{DAIR-V2X} \cite{yu2022dair} is the first real-world V2I collaborative perception dataset.
We set the LiDAR range as $x\in[-100.8m,100.8m],y\in[-40m,40m]$. 
The training set has 237,725 fully supervised objects and 9,622 sparsely supervised objects. 
\textbf{V2V4Real} \cite{xu2023v2v4real} is the first real-world V2V collaborative perception dataset.
The LiDAR range is set as $x\in[-140.8m,140.8m],y\in[-38.4m, 38.4m]$. The training set comprises 145,278 fully supervised objects and 14,210 sparsely supervised objects. 

\subsection{Experimental Setup}
\paragraph{Implementation Details}
We implement the CoDTS with OpenCOOD framework\cite{OPV2V}. Specifically, the collaborative detector is trained using the Adam optimizer with a learning rate of 0.002. The parameters for MFM, SFM, and NAS are set as follows: $\sigma^{st}_l=0.15$, $\sigma^{st}_h=0.2$,$\tau=0.15$, $\tau^{nei}=0.6$. For the EMA strategy, we set $\alpha=0.999$. The refinement starts iteration $I_{refine}$ is set to $0.5 \cdot I_{max}$. All models are trained on NVIDIA RTX A4000. 

\paragraph{Benchmarks}
We compare our CoDTS with sparsely supervised methods SSC3OD \cite{han2023ssc3od} and HINTED \cite{HINTED}. Our static teacher (ST) is the same as that of SSC3OD, which first applies the self-supervised learning method to pre-train the encoder. We then load the pre-trained encoder into ST and train it with sparse labels. To enable quantitative comparison, we utilize three classical intermediate collaborative detectors: F-Cooper \cite{fcooper}, AttFusion \cite{OPV2V}, and DiscoNet \cite{DiscoNet}. We keep the same supervised loss as these detectors used.

\subsection{Quantitative Evaluation}

\subsubsection{Comparison with Spasely Supervised Methods}
Table~\ref{table:performance} presents the collaboration detection results in both fully and sparsely supervised paradigms. The evaluation is conducted on four large-scale collaborative perception datasets. 
Due to insufficient supervision, collaborative detectors trained with sparse labels have lower accuracy than those trained with full labels. 
Specifically, SSC3OD and HINTED partially address this issue, but the performance of these detectors still significantly lags behind those trained with full labels. This is because these methods mainly rely on high-quality pseudo labels filtered by high confidence thresholds.

\begin{table}[htbp]
  \Huge
  \renewcommand\arraystretch{1.15} 
  \centering 
\resizebox{\linewidth}{!}{
\begin{tabular}{c|ccc|cc|cc}
  \toprule 
  \multirow{2}*{Detectors} & \multicolumn{3}{c}{Modules} & \multicolumn{2}{c}{Student $\theta^{s}$}   & \multicolumn{2}{c}{Dynamic Teacher $\theta^{dt}$}           \\
\cline{2-8}
  & MFM  & SFM  &  NAS & AP@0.5               & AP@0.7               & AP@0.5               & AP@0.7               \\
\midrule
\multirow{4}{*}{F-Cooper}   &  - & -&     -      & 49.49                & 32.58       & 49.49                & 32.58                \\
&   \checkmark & -&   -   & 66.90 & 42.92  & 67.90 & 47.33  \\
&   \checkmark & - &   \checkmark   & 66.86                & 47.11                & 68.55                & 48.72                \\
&   \checkmark & \checkmark&   \checkmark    & 68.73        & 48.05   & \bf{70.35}                & \bf{50.53}                \\
                           \midrule
\multirow{4}{*}{AttFusion} &  - & -&     -     & 51.21                & 32.93       & 51.21                & 32.93                \\
&   \checkmark & -&   -    &  65.58 & 48.77  & 67.25 &  50.35\\
&   \checkmark & - &   \checkmark     & 67.87     & 50.66     & 68.51                & 51.22                \\
&   \checkmark & \checkmark&   \checkmark      & 65.46                & 41.05     & \bf{69.23}                & \bf{51.24}                \\
\midrule
\multirow{4}{*}{DiscoNet}  &  - & -&     -     & 51.39                & 32.97      & 51.39                & 32.97                \\
&   \checkmark & -&   -   &  65.40 &  48.61  & 67.06 &  49.10 \\
&   \checkmark & - &   \checkmark   & 66.16                & 49.05     & 67.60                 & 49.77                \\
&   \checkmark & \checkmark&   \checkmark    & 67.27    & 45.73    & \bf{68.73}                & \bf{51.05}                \\
\bottomrule
\end{tabular}
}
\caption{Ablation study of main components on DAIR-V2X.}
\label{table:components}
\vspace{-0.5em}
\end{table}

Our CoDTS mines pseudo labels with both high-quality and high-quantity, enabling us to achieve the highest performance in sparsely supervised scenarios. Specifically, in the V2X-Sim, the detector trained with the CoDTS performs similarly to a fully supervised learning approach at IoU=0.3,0.5 and 0.7. In the OPV2V, our CoDTS improves the AP of collaborative detectors by over $10\%$.
In the DAIR-V2X and V2VReal, our CoDTS improves the AP of collaborative detectors by over $20\%$ at IoU=0.3 and 0.5, and by approximately $10\%$ at IoU=0.7. These results demonstrate the effectiveness of the CoDTS in generating pseudo labels.

\begin{table}[htbp]
  \Huge
  \renewcommand\arraystretch{1.2} 
  \centering 
\resizebox{1\linewidth}{!}{
  \begin{tabular}{cc|cc|cc|cc}
    \toprule
  \multicolumn{2}{c}{MFM}  & \multicolumn{2}{c}{F-Cooper}    & \multicolumn{2}{c}{AttFusion}   & \multicolumn{2}{c}{DiscoNet}    \\
  \midrule
  $\sigma^{st}_l$ & $\sigma^{st}_h$   & AP@0.5         & AP@0.7         & AP@0.5         & AP@0.7         & AP@0.5         & AP@0.7         \\
  \midrule
  0.1          & 0.2     & 69.61 & 49.80  & 68.83 & 51.02  & 68.52 & \textbf{51.23} \\
  0.1          & 0.25       & 69.03 & 49.22 & 68.72 & 51.31 & 67.84 & 49.33 \\
  0.1          & 0.3      & 67.93 & 48.92 & 67.37 & 50.11  & 67.86 & 49.26 \\
  \midrule 
  0.15          & 0.2    & 70.35 & \textbf{50.53} & 69.23 & \textbf{51.24}  & \textbf{68.73} & 51.05 \\
  0.15          & 0.25    & \textbf{70.43} & 50.02  & \textbf{69.56} & 51.10  & 67.77 & 50.49 \\
  0.15          & 0.3     & 70.13 & 49.92  & 68.92 & 50.74 & 67.51 & 50.25\\
  \bottomrule   
  \end{tabular}
}
\caption{Ablation study of $\sigma^{st}_l$ and $\sigma^{st}_h$ on DAIR-V2X.}
\label{table:threshold}
\vspace{-0.5em}
\end{table}

\subsubsection{Comparison with Semi-supervised Methods}
Given the limited research on sparsely supervised scenes and the fact that our proposed CoDTS employs a teacher-student framework similar to those used in semi-supervised methods, we also explore its performance in semi-supervised settings. Table~\ref{table:semi-supervised} presents a comparison of our results with various semi-supervised methods: 3DIoUMatch \cite{3DIoUMatch}, HSSDA \cite{HSSDA}, and HINTED \cite{HINTED}. 
Since it is difficult to pre-train collaborative detectors with a very low semi-supervised ratio, we randomly choose 10$\%$ and 20$\%$ of fully annotated frames from the training split.
Our method achieves the best performance in both 10$\%$ and 20$\%$ labeling rates, which demonstrates the superiority of our CoDTS in producing pseudo labels.

\subsection{Ablation Studies}

\subsubsection{Contribution of Main Components}
We conduct ablation studies to assess the effectiveness of the proposed MFM, SFM and NAS modules. We take collaborative detectors trained from scratch as a baseline and gradually incorporate each component. As shown in Table~\ref{table:components}, when we use MFM with a high confidence threshold and utilize the identified positive instances as pseudo labels, the performance of the student and dynamic teacher is improved. 

The performance is further enhanced when we combine MFM and NAS to generate pseudo labels, emphasizing the importance of sampling neighbor instances. By integrating MFM, SFM, and NAS, and implementing the staged training strategy, we achieve the best performance. This demonstrates the effectiveness of adaptive complementary learning in mining high-quality and high-quantity pseudo labels.
In the training process, since the dynamic teacher can be viewed as a temporal ensemble of students across different time steps, the dynamic teacher consistently outperforms the student. Furthermore, as employing varying confidence thresholds at different stages may impede the performance of the student, we choose the dynamic teacher for inference.

\subsubsection{Confidence Threshold for Static Teacher}
This section evaluates the influence of static teacher thresholds $\sigma^{st}_l$ and $\sigma^{st}_h$ on performance. Given that the OpenCOOD benchmark only considers predictions with scores above 0.2 during inference, and the maximum prediction score of the static teacher is below 0.35, we establish the low confidence threshold interval as (0, 0.2) and the high threshold interval as [0.2, 0.3]. 
We randomly select thresholds from the above intervals. Specifically, we set low confidence thresholds $\sigma^{st}_l$ at 0.1 and 0.15, and high confidence thresholds $\sigma^{st}_h$ at 0.2, 0.25, and 0.3. 
The results of different combinations of these confidence thresholds are in Table~\ref{table:threshold}. 

\begin{figure}[htbp]
  \centerline{\includegraphics[width=1\linewidth]{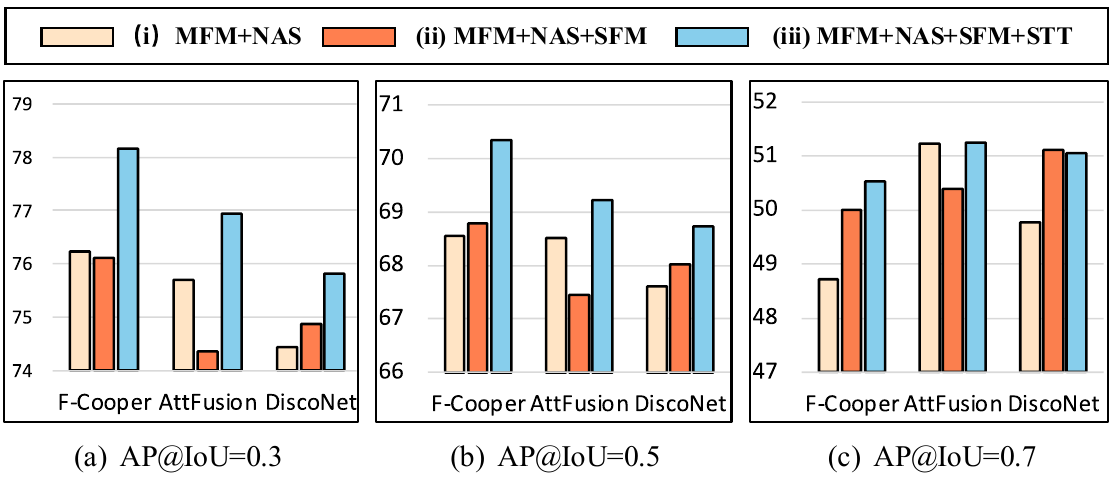}}
  \caption{Ablation study of staged training strategy (STT) on DAIR-V2X.  
  }
  \label{fig4}
  \vspace{-1.2em}
\end{figure}

The detection results across these combinations show no significant difference, with all achieving higher performance than the baseline methods in Table~\ref{table:performance}. This indicates that our framework does not rely on manually set low or high thresholds. As long as the MFM threshold falls within an appropriate range, our framework performs effectively.
Additionally, we tried to apply dynamic threshold adaptation to MFM but found the training to be unstable. This instability is due to the lower performance of the static teacher and its unstable predictions on sparse labels.

\subsubsection{The Importance of Staged Training Strategy}

We evaluate the importance of the staged training strategy (STT) and present the results in Figure~\ref{fig4}. The settings for the three methods are as follows: i) involves high confidence threshold-based MFM and NAS; ii) involves high confidence threshold-based MFM, NAS, and SFM. Specifically, we first train student and teacher models using a low-confidence threshold-based MFM and NAS. In the next stage, we take this teacher model as an auxiliary static teacher and apply SFM to identify missing instances. iii) involves MFM, NAS, and SFM, and utilizes the staged training strategy to achieve end-to-end training.

The results from (i) to (ii) indicate that directly employing SFM could improve collaborative detectors when IoU=0.7, since it provides high-quality missing positive instances for the MFM. However, its performance remains inferior to that of using STT, especially at IoU=0.3 and 0.5. 
This is because, two stages in (ii) are trained separately. In the second stage, the student is optimized from scratch, while the dynamic teacher remains static without further updates. 
In contrast, STT in (iii) enables end-to-end training, allowing the student to undergo continuous optimization. Additionally, the dynamic teacher is updated using the EMA of the student and generates partial labels to supervise the student, enabling their mutual learning. Therefore, the dynamic teacher can generate more complemented pseudo labels adaptively.

\section{Conclusion}
In this paper, we propose a collaborative perception dual teacher-student framework, CoDTS, which leverages adaptive complementary learning to generate pseudo labels in both high quality and high quantity. Specifically, the MFM generates high-quality pseudo labels, while the SFM adaptively identifies missing instances using a dynamic threshold. Furthermore, the NAS enhances the dense representation of pseudo labels during training. The framework employs a staged training strategy, where the warm-up stage pre-trains the student and dynamic teacher, and the refinement stage produces pseudo labels via adaptive complementary learning. Extensive experiments on four large-scale datasets demonstrate that CoDTS effectively addresses the imbalance of pseudo labels between quality and quantity.

\section{Acknowledgments}
This work was supported by the National Natural Science Foundation of China under Grant U2268203, 62203040 and 62306027.

\bibliography{references}

\section{Appendix}

\subsection{Problem formulation}
In a collaboration perception scenario involving $N$ agents, let $x_i$ and $y_i$ represent the LiDAR data and supervision of the $i$-th agent, respectively. Additionally, let $\xi_i$ denote the 6-DoF pose of the $i$-th agent. In the intermediate collaborative 3D detection, the ego vehicle receives and fuses features from other agents, which can be formulated as follows:
\begin{align}
  f_{i}&=\boldsymbol{F}_{\text{enc}}(x_{i}), \nonumber \\
  f_{j}&=\boldsymbol{F}_{\text{enc}}(x_{j}), \nonumber\\
  f_{j\rightarrow i}&=\boldsymbol{F}_{\text{proj}}(f_{j},(\xi_{i},\xi_{j})), \nonumber\\
  \label{eq1} \widetilde{f}_{i}&=\boldsymbol{F}_{\text{fuse}}(f_{i},f_{j\rightarrow i}), \nonumber\\
  z_{i}&=\boldsymbol{F}_{\text{det}}(\widetilde{f}_{i}), \nonumber
\end{align}
where $\boldsymbol{F}_{\text{enc}}$, $\boldsymbol{F}_{\text{proj}}$, $\boldsymbol{F}_{\text{fuse}} $ and $\boldsymbol{F}_{\text{det}}$ represent encoder, coordinate projection, feature fusion and detection head, respectively. We denote the fused feature as $\widetilde{f}_{i}$ and the detection output as $z_{i}$, which is supervised by detection loss $\mathcal{L}_{\text{det}} $.

In the sparsely supervised paradigm, each agent randomly annotates only one object. As shown in Table \ref{table:label number}, we count the number of frames in training sets for four large-scale collaborative perception datasets, along with the number of both full and sparse labels, and compute the proportion of sparse labels to all objects. The sparse subsets require annotation of only around $4\%$ to $10\%$ of objects, in contrast to the complete annotation of all objects in the original training set. 

\begin{table}[htbp]
  \Huge
  \renewcommand\arraystretch{1.3} 
  \centering 
  \resizebox{\linewidth}{!}{
  \begin{tabular}{cc|ccc}
  \toprule 
  Datasets & \# Train set &\# Full label & \# Sparse label & Sparse ratio \\
  \midrule
  V2V4Real \cite{xu2023v2v4real} &7,105 & 145,278 &	14,210	&$\textbf{9.78\%}$ \\
  DAIR-V2X \cite{yu2022dair} &4,811 & 237,725 &	9,622	&$\textbf{4.05\%}$ \\
  OPV2V \cite{OPV2V} &6,765&358,142&21,139&$\textbf{5.90\%}$ \\
  V2X-Sim \cite{V2X-Sim}& 8,000& 698,991 &	29,300	& $\textbf{4.19\%}$ \\
  \bottomrule
  \end{tabular}
  }
  \caption{Comparison of full and sparse annotation in collaborative detection.}
  \label{table:label number}
\end{table}

This limited supervision in sparsely supervised learning leads to a performance decrease in collaborative detectors. The traditional teacher-student frameworks only focus on the quality of pseudo labels and overlook the quantity. Our goal is to tackle this challenge and improve the performance of sparsely supervised collaborative detectors.

\subsection{Details of benchmarks}
This work focuses on sparsely supervised collaborative 3D object detection. Given that SSC3OD \cite{han2023ssc3od} is the only prior work on this specific task, we also consider benchmarks from sparsely supervised 3D object detection, such as SS3D \cite{liu2022ss3d}, Coln \cite{xia2023coin} and HINTED \cite{HINTED}. However, SS3D utilizes data augmentation to generate pseudo labels and stores them offline, which is unsuitable for intermediate collaboration in random collaborative environments. Besides,  Coln proposes multi-class contrastive learning to enhance the discriminability of features, However, most collaborative perception datasets have only one category, e.g., Car. Therefore, the relevant benchmarks for sparsely supervised collaborative 3D object detection are SSC3OD and HINTED. 
The main differences among SSC3OD, HINTED, and CoDTS as plugins for generating pseudo labels are as follows:

\textbf{SSC3OD} includes a static teacher and a student. It focuses on self-supervised learning and proposes a mask autoencoder-based method to pre-train the teacher model, then uses this static teacher to generate high-quality pseudo labels by applying a high threshold.

\textbf{HINTED} includes a dynamic teacher and a student. It focuses on mining hard instances by employing a mixed-density student, which involves inputting both original and downsampled LiDAR data into the student model. The pseudo labels are then used to supervise the original, downsampled, and mixed density features, respectively. The dynamic teacher filters pseudo labels with high and mid-confidence thresholds, assigning a low weight to the mid-confidence pseudo labels.

\textbf{CoDTS} includes a static teacher, a dynamic teacher, and a student. CoDTS leverages adaptive complementary learning to generate pseudo labels with both high quality and quantity through a staged training strategy. During the warm-up stage, we employ a static teacher with a low threshold to train the dynamic teacher, ensuring that the dynamic teacher can effectively detect most objects. In the refinement stage, the static teacher utilizes a high threshold to produce high-quality pseudo labels, while the dynamic teacher identifies missing instances using a high confidence threshold, which ensures both the quality and quantity of pseudo labels.

\subsection{Ablation study on staged training strategy} 
We evaluate the importance of the staged training strategy (STT) and present the results in Table~\ref{tab2}-Table~\ref{tab4}. Each table includes the following information: i) the first row displays the performance of high-confidence threshold-based MFM and NAS; ii) the second and fourth rows incorporate high-confidence threshold-based MFM, NAS and SFM. Initially, we train student and teacher models using a low-confidence threshold-based MFM and NAS. Subsequently, we leverage the teacher model as a static auxiliary teacher and apply SFM to identify missing instances. iii) The third and fifth rows present the results of the CoDTS framework, incorporating MFM, SFM, and NAS, along with the staged training strategy for end-to-end learning.

The results indicate that directly employing SFM could improve collaborative detectors since it provides high-quality missing instances for the MFM. However, its performance remains inferior to that of using STT. This is because STT enables the model to be trained by pseudo labels with different quality and quantity, ensuring continuous optimization and a more comprehensive understanding of detection knowledge. Furthermore, by implementing STT, the student and dynamic teacher models are optimized through mutual learning, allowing them to evolve together and continuously enhance detection accuracy. Therefore, the staged training strategy is essential for the CoDTS.

\subsection{The importance of warm-up stage}
The proposed CoDTS adopts a staged training strategy. 
While the framework can be trained using only the refinement stage, this approach leads to reduced detection accuracy compared to the full CoDTS training process.
This is because dynamic teacher lacks strong detection capability at first, so applying SFM directly introduces substantial noise, reducing model accuracy. In contrast, the warm-up stage in CoDTS initializes dynamic teacher and improves its detection capability, ensuring high-quality pseudo labels are produced by SFM in the refinement stage.

\begin{table}[htbp]
  \large
  \renewcommand\arraystretch{1.2} 
  \centering 
  \resizebox{0.8\linewidth}{!}{
  \begin{tabular}{c|ccc}
  \toprule 
  F-Cooper + CoDTS & AP@0.3 &AP@0.5 &AP@0.7  \\
  \midrule
  w/o warm-up  & 68.08 & 61.48 &	42.16	 \\
  w/ warm-up  & 78.17 & 70.35 & 50.53 \\
  \bottomrule
  \end{tabular}
  }
  \caption{The importance of warm-up stage.}
  \label{table:wam-up stage}
  \vspace{-1em}
\end{table}

\subsection{Number of pseudo-labels by different methods}

We conduct experiments on the real-world dataset DAIR-V2X \cite{yu2022dair} (4,811 training scenes) to count the average number (AN) of pseudo labels in each frame, and calculate the false prediction ratio (FPR) and missing prediction ratio (MPR).

Table~\ref{table:pseudo label number} shows that CoDTS produces significantly more pseudo labels than existing methods, with additional pseudo labels precisely covering areas where previous methods missed. Therefore, our pseudo labels have lower FPR and MPR, ensuring both high quality and high quantity.

\begin{table}[htbp]
  \scriptsize
  \renewcommand\arraystretch{0.9} 
  \centering 
  \resizebox{0.8\linewidth}{!}{
  \begin{tabular}{c|ccc}
  \toprule 
  F-Cooper based & AN & FPR & MPR  \\
  \midrule
  SSC3OD  & 3.84 & 0.21 &	0.75	 \\
  HINTED  & 6.48 & 0.23 &	0.66	 \\
  CoDTS  & 12.33 & 0.23 & 0.49 \\
  \bottomrule
  \end{tabular}
  }
  \caption{Comparison of pseudo label number.}
  \label{table:pseudo label number}
  \vspace{-1em}
\end{table}

\subsection{Visualization of detection results} 
To demonstrate the effectiveness of the CoDTS, we display the visualization results of the sparsely supervised collaborative detectors AttFusion on four datasets, as shown in Figure~\ref{fig1}-Figure~\ref{fig4}.
Compared to collaborative detectors trained with SSC3OD and HINTED, the detector trained with CoDTS consistently achieves superior performance. Specifically, our CoDTS is capable of adapting to different LiDAR distributions and densities across both simulated and real-world datasets. Besides, the CoDTS can effectively detect instances in sparse point clouds and distant areas, or those that are closely adjacent.

\begin{figure}[htbp]
  \centerline{\includegraphics[width=1\linewidth]{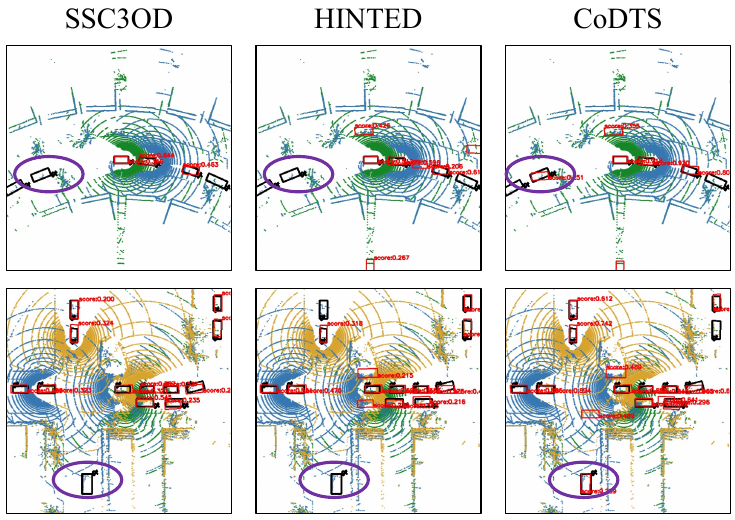}}
  \caption{Qualitative comparison on V2X-Sim. Black and \textcolor[RGB]{255,0,0}{red} boxes denote the ground truths and detection results.}
  \label{fig1}
  \vspace{-1em}
\end{figure}

\begin{figure}[htbp]
  \centerline{\includegraphics[width=1\linewidth]{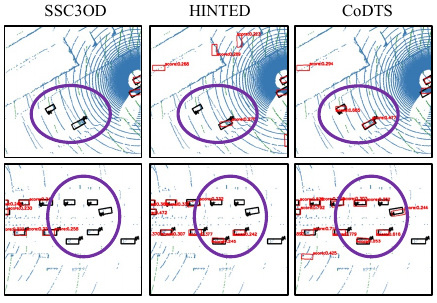}}
  \caption{Qualitative comparison on OPV2V. Black and \textcolor[RGB]{255,0,0}{red} boxes denote the ground truths and detection results.}
  \label{fig2}
  \vspace{-1em}
\end{figure}

\begin{figure}[htbp]
  \centerline{\includegraphics[width=1\linewidth]{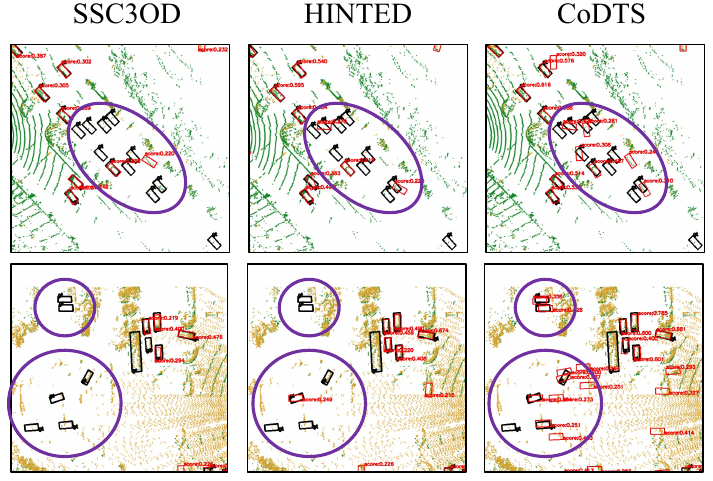}}
  \caption{Qualitative comparison on DAIR-V2X. Black and \textcolor[RGB]{255,0,0}{red} boxes denote the ground truths and detection results.}
  \label{fig3}
  \vspace{-1em}
\end{figure}

\begin{figure*}[htbp]
  \centerline{\includegraphics[width=0.9\linewidth]{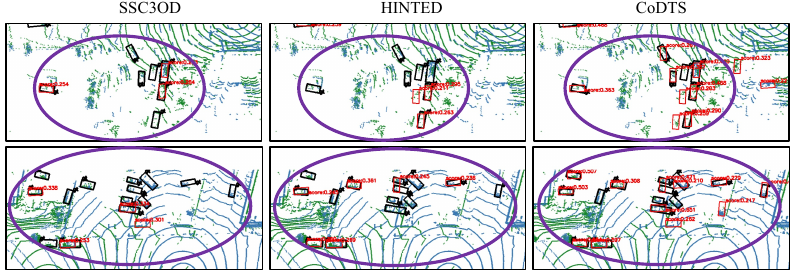}}
  \caption{Qualitative comparison on V2V4Real. Black and \textcolor[RGB]{255,0,0}{red} boxes denote the ground truths and detection results.}
  \label{fig4}
  \vspace{-1em}
\end{figure*}

\begin{table*}[htbp]
    \centering
  \renewcommand\arraystretch{1.1} 
  \resizebox{0.9\linewidth}{!}{
  \begin{tabular}{ccc|ccc|ccc|ccc}
    \toprule
    \multirow{2}*{$\sigma^{st}_l$} & \multirow{2}*{$\sigma^{st}_h$} & \multirow{2}*{STT}     & \multicolumn{3}{c}{F-Cooper}                     & \multicolumn{3}{c}{AttFusion}                    & \multicolumn{3}{c}{DiscoNet}                     \\
    \cline{4-12}
      &   &                      & AP@0.3         & AP@0.5         & AP@0.7         & AP@0.3         & AP@0.5         & AP@0.7         & AP@0.3         & AP@0.5         & AP@0.7         \\
    \midrule
     -          & 0.2          &  -                    & 76.23          & 68.55          & 48.72          & 75.70           & 68.51          & 51.22          & 74.45          & 67.60           & 49.77          \\
     \midrule
    0.1        & 0.2          & - & 76.59          & 68.87          & 49.12          & 74.12          & 67.18          & 49.76          & 73.56          & 66.88          & 49.24          \\
    0.1        & 0.2          & \checkmark     & \textbf{77.67} & \textbf{69.61} & \textbf{49.80}  & \textbf{76.38} & \textbf{68.83} & \textbf{51.02} & \textbf{75.72} & \textbf{68.52} & \textbf{51.23} \\
    \midrule
    0.15       & 0.2          &   -       & 76.12          & 68.79          & 50.01          & 74.36          & 67.46          & 50.40           & 74.87          & 68.02          & 50.02          \\
    0.15       & 0.2         &   \checkmark       & \textbf{78.17} & \textbf{70.35} & \textbf{50.53} & \textbf{76.95} & \textbf{69.23} & \textbf{51.24} & \textbf{75.82} & \textbf{68.73} & \textbf{51.05} \\
    \bottomrule
    \end{tabular}
  }
  \caption{Ablation study of staged training strategy on DAIR-V2X, where $\sigma^{st}_h=0.2$.}
    \label{tab2}
    \vspace{-1em}
\end{table*}

\begin{table*}[htbp]
  \centering
  \renewcommand\arraystretch{1.1} 
  \resizebox{0.9\linewidth}{!}{
  \begin{tabular}{ccc|ccc|ccc|ccc}
  \toprule
  \multirow{2}*{$\sigma^{st}_l$} & \multirow{2}*{$\sigma^{st}_h$} & \multirow{2}*{STT}     & \multicolumn{3}{c}{F-Cooper}                     & \multicolumn{3}{c}{AttFusion}  & \multicolumn{3}{c}{DiscoNet}                     \\
    \cline{4-12}
      &   &                      & AP@0.3         & AP@0.5         & AP@0.7         & AP@0.3         & AP@0.5         & AP@0.7         & AP@0.3         & AP@0.5         & AP@0.7         \\
  \midrule
      -    & 0.25          &        -             & 70.40          & 63.53          & 45.29          & 68.88          & 62.74          & 45.64          & 68.67          & 60.81          & 39.86          \\
             \midrule
  0.1        & 0.25          & -  & 76.24          & 68.09          & 45.79          & 72.50          & 65.20          & 45.94          & 70.35          & 64.46          & 45.31          \\
  0.1        & 0.25        & \checkmark                   & \textbf{76.41} & \textbf{69.03} & \textbf{49.22} & \textbf{75.55} & \textbf{68.72} & \textbf{51.31} & \textbf{74.81} & \textbf{67.84} & \textbf{49.33} \\
  \midrule
  0.15       & 0.25        &    -             & 74.87          & 67.84          & 49.38          &     72.50     &    66.48        &     50.81        & 72.36          & 65.62          & 47.72          \\
  0.15       & 0.25        & \checkmark     & \textbf{78.77} & \textbf{70.43} & \textbf{50.02} & \textbf{76.96} & \textbf{69.56} & \textbf{51.10}  & \textbf{74.42} & \textbf{67.77} & \textbf{50.49} \\
  \bottomrule
  \end{tabular}
  }
    \caption{Ablation study of staged training strategy on DAIR-V2X, where $\sigma^{st}_h=0.25$.}
    \label{tab3}
    \vspace{-1em}
\end{table*}

\begin{table*}[htbp]
  \centering
  \renewcommand\arraystretch{1.1} 
  \resizebox{0.9\linewidth}{!}{
  \begin{tabular}{ccc|ccc|ccc|ccc}
    \toprule
  \multirow{2}*{$\sigma^{st}_l$} & \multirow{2}*{$\sigma^{st}_h$} & \multirow{2}*{STT}     & \multicolumn{3}{c}{F-Cooper}                     & \multicolumn{3}{c}{AttFusion}                    & \multicolumn{3}{c}{DiscoNet}                     \\
    \cline{4-12}
      &   &                      & AP@0.3         & AP@0.5         & AP@0.7         & AP@0.3         & AP@0.5         & AP@0.7         & AP@0.3         & AP@0.5         & AP@0.7         \\
  \midrule
      -       & 0.3          &  -   & 63.98          & 58.41          & 38.35          & 65.19          & 58.27          & 32.48          & 64.53          & 59.59          & 42.28          \\
             \midrule
  0.1        & 0.3           &  -   & 75.48          & 67.61          & 44.98          &     67.12        &     62.35           &       47.00      & 67.01          & 62.37          & 46.74          \\
  0.1        & 0.3           & \checkmark   & \textbf{75.30}  & \textbf{67.93} & \textbf{48.92} & \textbf{74.16} & \textbf{67.37} & \textbf{50.11} & \textbf{74.81} & \textbf{67.86} & \textbf{49.26} \\
  \midrule
  0.15       & 0.3           &  -   & 73.65          & 66.91          & 49.60          &   72.28         &    64.99      &     46.31      & 72.50          & 64.78          & 43.87          \\
  0.15       & 0.3           & \checkmark   & \textbf{78.38} & \textbf{70.13} & \textbf{49.92} & \textbf{76.48} & \textbf{68.92} & \textbf{50.74} & \textbf{74.31} & \textbf{67.51} & \textbf{50.25}\\
  \bottomrule
  \end{tabular}
  }
  \caption{Ablation study of staged training strategy on DAIR-V2X, where $\sigma^{st}_h=0.3$.}
    \label{tab4}
    \vspace{-1em}
\end{table*}

\end{document}